\documentclass{article}

\usepackage{arxiv}

\usepackage[utf8]{inputenc} 
\usepackage[T1]{fontenc}    
\usepackage{hyperref}       
\usepackage{url}            
\usepackage{booktabs}       
\usepackage{amsfonts}       
\usepackage{nicefrac}       
\usepackage{microtype}      
\usepackage{lipsum}
\usepackage{graphicx}
\graphicspath{ {./images/} }

\title{LPar - A Distributed Multi Agent platform for building Polyglot, Omni Channel and Industrial grade Natural Language Interfaces }

\author{
 Pranav Sharma \\
  Cognizant Worldwide Limited\\
  1 Kingdom St\\
   London, W2 6BD \\
  \texttt{pranav.sharma@cognizant.com} }

\begin{document}
\maketitle
\begin{abstract}
The goal of serving and delighting customers in a personal and near human like manner is very high on automation agendas of most Enterprises. Last few years, have seen huge progress in Natural Language Processing domain which has led to deployments of conversational agents in many enterprises. Most of the current industrial deployments tend to use Monolithic Single Agent designs that model the entire knowledge and skill of the Domain. While this approach is one of the fastest to market, the monolithic design makes it very hard to scale beyond a point. There are also challenges in seamlessly leveraging many tools offered by sub fields of Natural Language Processing and Information Retrieval in a single solution. The sub fields  that can be leveraged to provide relevant information are, Question and Answer system, Abstractive Summarization, Semantic Search, Knowledge Graph etc. Current deployments also tend to be very dependent on the underlying Conversational AI platform (open source or commercial) , which is a challenge as this is a fast evolving space and no one platform can be considered future proof even in medium term of 3-4 years. 
Lately,there is also work done to build multi agent solutions that tend to leverage a concept of master agent. While this has shown promise, this approach still makes the master agent in itself difficult to scale. 
To address these challenges, we introduce LPar, a distributed multi agent platform for large scale industrial deployment of polyglot, diverse and inter-operable agents. The asynchronous design of LPar supports dynamically expandable domain. We also introduce multiple strategies available in the LPar system to elect the most suitable agent to service a customer query. 

\end{abstract}


\section{Introduction}
Natural Language Queries from customers of any large Enterprise are very diverse and complex. A lot of effort to model and automate these in last few years have been as monolithic applications leveraging Conversational AI tools available in market or available Open source. 

While these efforts have given encouraging results but have also highlighted numerous challenges. 

Following are some of the key challenges  -

\begin{itemize}

\item NLP is a constantly evolving space and months of development work done to build and deploy these Agents very quickly become out of date as new Research or tools come into market. 
\item Monolithic nature of design of these Agents makes the development slow and very difficult to federate, leading into increased time to model the depth and breadth of relevant domain of the enterprise.
\item Addition of New Skills to the Agent, along with build effort, also requires extensive regression testing effort as training data of new skill can very easily impact already existing skills. 
\item Queries that require automation vary vastly in terms of number of turns and data points/slots required from user to provide a resolution with optimal experience. Some factoid based queries can be resolved in a single pass while other goal oriented skills require complex multi turn conversation. Also there are some queries tend to be more suitable to a search domain. Efficient coverage of the depth and breadth of the domain requires more that just one tool to model the queries.
\item The interpretation of user queries tend to be very literal basis what a user said/texted. To have a meaningful communication, it is very important to interpret what user said basis deep understanding of the user and his/her current context. 

\end{itemize}

With LPar, we intend to resolve many of above challenges and provide an  framework that provides stability and consistency to large Organizations. 
\subsection{Tools for building Conversational Systems}

As mentioned above, Customer queries are very diverse and it’s always a challenge and no one tool can be used to model them all. Based on our experience and research we following tools currently available 

\paragraph{Goal Oriented Agents}

Goal Oriented Bots use Natural Language Understanding, dialog management and response generation to fulfill goal(s) for the user of the system. These tend to gather user inputs (Slots) in multi turn conversations and retain context across turns. Once a user’s intent and inputs have been confirmed, these perform a fulfilment function leveraging APIs. Last few years have seen increased adoption of these in the industry and are used to automate standard user interactions like checking flight status, flight bookings, checking balances etc. With the current commercial and open sources tools in market, these require conversation modelling and build. 

\paragraph{FAQ Agents }

FAQ bots generate contextual vectors for large number of question and answer pairs  and indexes them in a persistence store. They usually serve user queries in a single pass. The user queries at the run time are transformed into corresponding contextual vector and a Nearest Neighbour search is performed on the persistent store using various distance measures like cosine similarity, Euclidean Distance , Manhattan Distance etc.  These are a great tool to cover a large breadth of queries with in the domain very quickly but usually lack the multi turn conversations and hence are not suitable for interactive queries. They also require a content cu-ration, review  and management effort from the SMEs to identify and maintain relevant Question and Answer pairs. 

\paragraph{Question and Answer Agents}

Question And Answer systems intend to provide a concise response to a natural language query from the user over a large knowledge Base. This task requires natural language  understanding of user query, Knowledge representation and reasoning over a domain. A question and answer system apart from providing the most relevant document, also looks to provide the span of the answer with in the document. 

\paragraph{Semantic Search Agents }

Semantic Search systems look to leverage the advances in Natural Language Processing to make search queries more relevant. Multiple tasks in NLP domain like embeddings, Entity Extraction, Sentiment Analysis , Named Entity Recognition are performed on content during the time of indexing to make the content more easily search-able during the query time. Context Vectors are created for user queries and nearest neighbour search is performed during the query time to find most relevant matches. Semantic search is usually a good fallback option for conversational system as they increases the domain coverage of the system without requiring development effort for every topic. It can also be a handy tool to reduce the expensive human hand over costs as some of the answers can be provided to the user by search itself.

\paragraph{Knowledge Graphs}

Knowledge Graphs are a powerful tool to represent interconnections of the world. They are used to store entities, relationships and attributes. They have been extensively utilised to model the knowledge of the domain and make the information more easily search-able through its interconnections. 
They have been used to build internet scale search systems and virtual agents.

\section{Design Goals}

Based on the review of above available tools and the challenges faced in current implementations, following are the key design goals for LPar

\paragraph{Multiple Application Support}
LPar envisions to provide support for multiple use cases from a single deployment. This will allow enterprises to scale the solution very quickly. 
\paragraph{Omni Channel }
Lpar supports most prevalent channels of communication that are required my modern enterprises. These include  Digital (web and mobile), Social Media , email , SMS and IVR. 

\paragraph{Moving from Monolith Centralised Agents to Distributed Micro Agents}

One of the key design goal for LPar is to have distributed Micro Agents that specialise in one task or topic. This allows Conversational systems to be more agile, flexible and maintainable. The design principles of leveraging micro components has already proven with in the wider software industry. 

\paragraph{Support for Multiple Tools (Conversational AI , Question and Answer Systems , FAQ Automation , Semantic Search , Human Agents)}

As described in the introduction section, there are multiple tools available for model natural language queries of users of a Conversational System. We were keen to enable designer and developers of conversational systems to use  use most suitable tool to model  user queries . The framework  intends to provide seamless switching across tools.

\paragraph{Polyglot and Technology Agnostic}

There are numerous Open Source and Commercial tools available in market for building conversational systems. With a fast moving technology domain the shelf life of any of these systems is very hard to predict.  The framework intends to provide an adapter based design pattern to maintain loose and temporal coupling with these tools. This will enable enterprises to very quickly wholly or partially migrate to newer, more robust tools as they come.   

\paragraph{Dynamically Expandable Domain}
Like to any machine learning system, agility and frequent updates is important to any conversational system. LPar aims to provide capability to add/update skills without brining down the overall system. 

\paragraph{Policy driven Disambiguation}

In a multi agent setup, there will be scenarios where more than one serving bot may be able to answer a query. It is important to have a policy driven approach to Disambiguation. Currently LPar supports 3 different policies . We expect the Policies to continue to evolve in future.

\paragraph{Utilize Internal and External Context}

A user of conversational system has a context from historical sessions and current session. To make conversations more natural, systems need to effectively utilise this information to have a more engaging conversation with the user. The context can also be leveraged to predict the next best action for the user. Most large Enterprises, capture a customers various interactions across their application landscape. As users engage with applications, they generate interaction data which can be effectively harnessed by conversational systems to improve the experience of the users. LPar is envisioned to get External context from these systems of interactions and leverage the same to improve customer experiences.  

\paragraph{Experimentation in production and Offline}
Natural language processing is a fast evolving space. We were keen to provide native support in the framework for fast experiments with minimal impact on users of the system.

\paragraph{Adaptive Dialog Management through Hyper Personalization}
To deliver optimal user experience,  LPar endeavours to understand its user’s context by integrating with Enterprise Context Service and personalise user’s journey.

\newpage

\begin{figure} 
\section{LPar - System Architecture}
Below is the high level diagram representing various components of LPar.

\bigskip
\newblock
    \centering
    \includegraphics [keepaspectratio=true,scale=0.4] {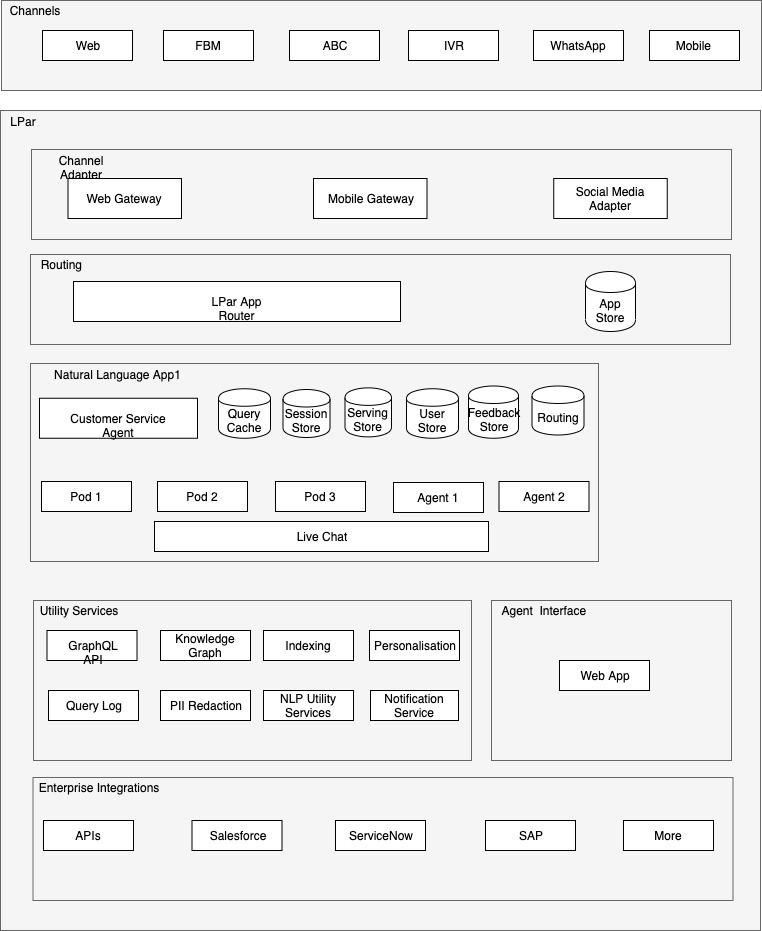} 
\end{figure}

\subsection{App Store }

LPar platform is designed to support multiple Natural Language Apps from a single deployment.  App store retains metadata about for all Natural Language applications supported by the system. These applications ideally would support different domains across the enterprise.

\begin{itemize}
  \item Name of Application
  \item Application Identifier
  \item Identifiers of Channels Supported
  \item Serving Matrix
  \item Resilience Rating for the application
  \item Data Classification Rating of Conversational Application
\end{itemize}
\newpage

\begin{figure} 

\subsection{LPar App Router}

The LPar App router routers the requests to appropriate Natural Language Apps with in the system basis the Application Identifier that is generated by the App store.

\subsection{Natural Language Apps}
Natural Language Apps are logically separate applications that support a business function. A single Natural Language app can comprise of multiple Agents and PODs.

\newblock

    \centering
    \includegraphics [keepaspectratio=true,scale=0.6] {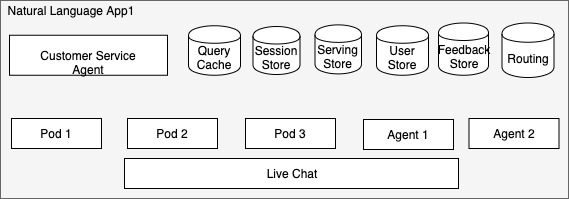} 
\end{figure}

The Organization design of the Natural Language App is crucial to overall quality of the service the app can deliver. The platform supports for both Hierarchical and Flat Organisational structures basis the requirement of a domain. For some enterprises, there may be a merit for the Natural Language App Organization Design to resemble closely to the structure of their respective areas. This can allow for Federation and parallelisation of the Natural Language App build and support to the most relevant teams rather than in a central structure.

\subsubsection{Customer Service Agent}

Customer Service Agent is responsible for overall experience of the user. 
This component performs following functions 

\begin{itemize}
  \item Act as Front line agent that receives user's request, understands user's context and recommends the most suitable Agent with in the system to service the request
  \item Executes the Agent Selection Process leveraging Broadcast Only and Search and Multicast strategies available in the LPar system 
  \item Responsible for establishing user session and to maintain context of the user as the conversation progresses 
  \item Leverage the Global User Context Service to personalise customer journey. Also, update the local context into the Global User Context Service.    
  \item Perform specialised routing as defined in the routing profile 
\end{itemize}

The key point to note is that the main function of the Customer Service Agent is to try and innately understand the user of the system. It is not a master agent that will have the knowledge of the entire domain.

\newpage

\subsubsection{Key Data Stores}

Natural Language Apps utilizes a number of data stores to service user queries.  The data stores are separate for all every Natural language App on the platform to support data privacy and segregation requirements. 

Following is the brief description of each of the data store 

\subsubsubsection{Session Store}

It maintains the user session and context as conversations progress. Customer Service Agent component utilizes and maintains this store to create and update user sessions. Every user session has an Serving Agent field that specifies the Agent with which the conversation is currently progressing. Customer Service Agent forwards all user queries to Serving Agent's  private address and processes its response. When the Serving Agent is not able to service a user's query and replies with an out of scope response, agent selection process is triggered as described in later sections. Session store also maintains Context which is Intents and Entities as provided by Serving Agents in their responses. In one Session a customer can be serviced by multiple serving agents basis the query from the user. 

\subsubsubsection{Serving Store}

It stores the list of available Pods and Agents which are available with in the system and can be directly reached by the Customer Service Agent. Following are they key attributes that are maintained by Serving Store 

\begin{itemize}

\item Agent Identifier
\item Agent Name
\item Agent Version 
\item Agent Type (POD or Agent)
\item Centroid of vectors of utterances used to train the agents
\item Connection protocol 
\item Private Address (Adapter Request and Response topics)
\item Status (Online or Offline)
\item Scope (Internal or External)
\item Class (Conversational Bot, FAQ bot, Q and A system, Semantic Search ,  Knowledge Graph, NA) 
\item Agent Rating ( Beginner, Intermediate, Professional, Expert)
\item Channels Supported
\item Average Response time 
\end{itemize}

\subsubsubsection{User Store}
It maintains the User profiles for the system. LPar system can also work with external user profile stores like CRM systems. User Profiles store users identifies across different channels so than an Omni Channel experience can be provided to the user. They can start a conversation in one channel and continue the same in a different channel. The attributes from User Profile store are key for the Customer Service Agent to provide a personalized experience to user. Attributes from User Profile are leveraged to choose the Customer Service Agent Persona that is most suitable for the customer. 

\subsubsubsection{Feedback Store}
It logs the feedback that users provide through various surveys that are run by the system. This the feedback and other parameters are utilized by the system is to update the Agent Rating. Agent Rating is leveraged by the system in Response Selection Policies. These become relevant when Multiple Agents provide responses to a users query

\subsubsubsection{QueryCache}

All incoming user queries in the system are logged in the query cache by the Customer Service Agent. It also records the response provided by the Agent. When Agent Selection Process is executed, QueryCache is leveraged to log responses from multiple agents and finally most suitable response is selected. 

\subsubsubsection{Routing}

Routing store is used by the system to store any recommended routing options for the user basis the users interactions and transactions across the enterprise

\subsubsection{Anatomy of a POD}

Pod is a core construct of Lpar that brings together a capability around a high level topic for the Domain of the Application. A POD can comprise of multiple Agents or PODs as well giving scalability to expand the domain of Natural Language application in both depth and breadth. Agents and PODs can be dynamically added or removed at the run time without any impact to overall Application. A Pod can have Agents of same or multiple different Classes (Conversational Bot, FAQ bot, Q and A system, Semantic Search ,  Knowledge Graph). A suggested way to model a domain would be to have a healthy mix of Agents belonging to different Agent Classes basis the requirements of the the domain.

The Pod Co-ordinator is responsible to find the most suitable Agent and provide its response for a particular user query leveraging all the members of the POD listed in the Pod member store.

\noindent%
\begin{minipage}{\linewidth}
\makebox[\linewidth]{
  \includegraphics[keepaspectratio=true,scale=0.4]{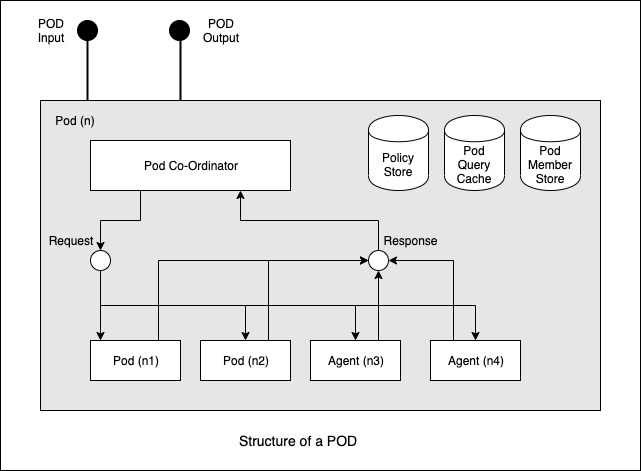}} 
  
\end{minipage}

\paragraph{Pod Addresses}

\begin{itemize}
  \item  Pods subscribe to Broadcast Request topic and publish processed response  back to Broadcast Response topic
  \item Pods can  dynamically subscribe and publish back to multicast topics during the run time 
  \item Pods also have individual request and response topics through which they can be contacted once a sessions has been established with them
\end{itemize}

Following are the examples of Agents that belong the Payments Pod

\begin{itemize}
  \item Bill Payments Agent  (Goal Oriented Agent)
  \item Direct Debit Agent  (Goal Oriented Agent)
  \item Payment Status Agent (Goal Oriented Agent)
  \item International Payments Agent (Goal Oriented Agent)
  \item Payments FAQ Agent  ( FAQ Agent)
  \item Payment Terms and Condition Agent (Semantic Search Agent)
\end{itemize}

\subsubsection{Anatomy of an Agent}

An Agent is the atomic unit of the Natural Language App bringing in a focused execution or query capability to the Application. A typical agent will perform Natural Language understanding to understand the user’s intent, extract entities from user’s input, perform validations, manage the dialog , execute an action and provide a response back to the user.  

Every agent integrates with an Adapter to keep the over all Application vendor agnostic. LPar system supports development of Agents using any commercial or open source solution. Multiple agents built on different technologies can co-exist with in a single Natural Language Application. 

\paragraph{Agent Addresses}

Agents have three key addresses using which they can be passed a message for a response. 
\begin{itemize}
  \item  Agents subscribe to Broadcast Request topic and publish processed response  back to Broadcast Response topic
  \item Agents can  dynamically subscribe and publish back to multicast topics during the run time 
  \item Agents also have individual request and response topics through which they can be contacted once a sessions has been established with them
\end{itemize}

\bigskip

\noindent%
\begin{minipage}{\linewidth}
\makebox[\linewidth]{
  \includegraphics[keepaspectratio=true,scale=0.4]{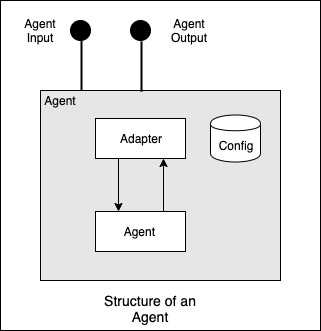}} 
  
\end{minipage}

\subsubsection{Agent Selection Process}

Customer Service Agent (at Application level) and Pod-Coordinator (at Pod level)  are the two components that have the responsibility to run the Agent Selection Process. There are two strategies for Agent selection namely \textbf{Broadcast only} and \textbf{Search and Broadcast}. The Natural Language App developers are expected to identify the most suitable strategy during the build time and configure the same in the App store settings.  

\newpage

Following flowchart depicts the agent selection process

\bigskip

\noindent%
\begin{minipage}{\linewidth}
\makebox[\linewidth]{
  \includegraphics[keepaspectratio=true,scale=0.4]{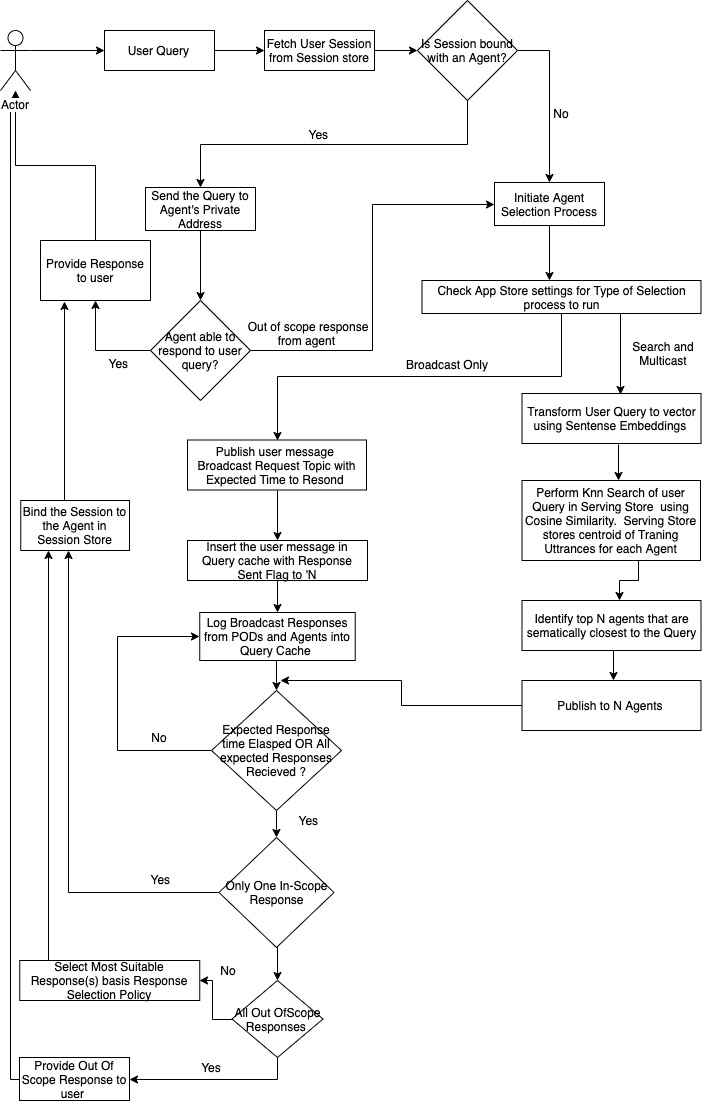}} 
  
\end{minipage}

\bigskip

Once the Agent selection process is complete and an agent has been identified, the user session is bound to the selected agent. Subsequent queries are sent directly to the Agent. If the agent is able to fulfil the query then the dialog continues with the Agent. If at some point the the agent is not able to provide response to the user query and provides an out of scope response, this leads to the Agent selection process getting called again to identify the next agent.

\subsubsubsection {Response Selection Policy}

\subsection{Live Chat and Human Agent}

Most Enterprise platform require capability to hand over to human agent under multiple different scenarios. Following are the standard key scenarios that require Human Agent Hand Over and are supported by LPar System 

\begin{itemize}
  \item User Explicitly asking to talk to a human
  \item A customer's sentiment drops below a threshold 
  \item Multiple Out OF Scope responses from the Agents with in the system
\end{itemize}
\newpage

LPar system provide connectors to multiple popular Live Chat platforms. 

\subsection{Utility Services}

LPar system provides multiple utility services to the App developers to support the App development effort. These are typically reusable services that are required by multiple applications. 

Following is the brief about each of the service 

PII Redaction - The service users regular experession and machine learning algorithms to redact PII information.

Profanity Filter - Identify if users are using inappropriate language and redact the same.

Sentiment Analysis - Provide the Sentiment of user queries.

Notification Service - Provides capability to send Push Notification to the users to enhance the experience.

Personalisation or Context Service - Provides Real time Context information for the user from across the enterprise that is leveraged by the platform to take simplify customer experience.

GraphQL APIs - These profile a unified schema for the Agents to utilize to support the integration with rest of the enterprise

Validation APIs - Validation APIs provide a common framework of validation across the Agents

\subsection{Channel Adapter}

Channel Adapters provide multiple connectivity options for the platform to integrate with Digital, Social Media and Voice channels. They abstract the channel connectivity details from rest of the platform.

\section{Experimental Setup}

Retail Banking domain was chosen for creating an experimental setup. 
Following are the multiple agents that were created in the system

\begin{itemize}
  \item Balance and Transaction Agent was created to have the responsibility to show customer balance for multiple accounts
  \item Product Finder Agent was created to provide answers for Product related Queries   
  \item Branch and ATM Finder Agent was created to provide responses about Branch and ATM locations 
  \item Payments Agent was created to showcase payment execution capability
  \item Connect Agent was created to provide the human handover capability.
  
  \end{itemize}

Multi channel capability was added to the system with Adapters for Facebook Messenger, WhatsApp and Amazon Connect (For Voice). 

Along with core intents to support the goal that each Agent was designed for, we also created an IndentifyYourself Intent in all Agents. On invocation of this intent with user utterances like "who are you" the Agent responded with its name.

We fired multiple different utterances in no specific order to try the capability of the system to seamlessly assign most relevant Agent to user session and provide quick responses to the customer. We also tried utilizing data that was provided by user to one Agent into another agent to verify the context retention capability of the system. We could demonstrate through our experimental setup LPar system is able to assign most relevant agent to support different user queries. The response times were also with in the acceptable limits.

\bigskip

\noindent%
\begin{minipage}{\linewidth}
\makebox[\linewidth]{
  \includegraphics[keepaspectratio=true,scale=0.4]{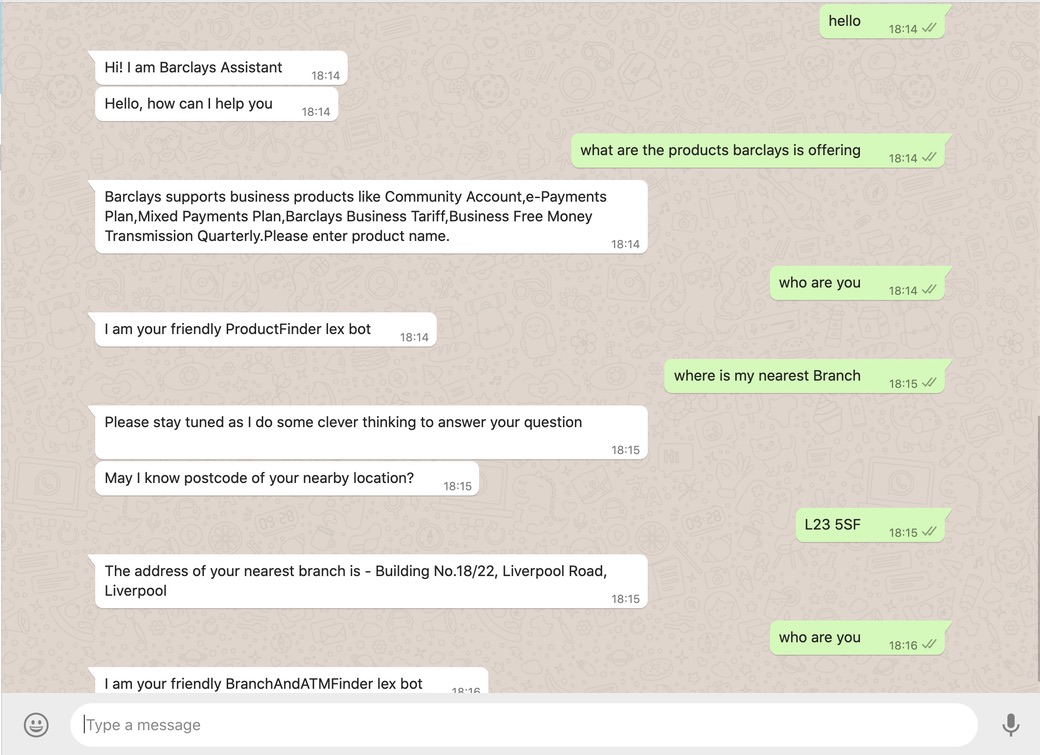}} 
 
\end{minipage}

\bigskip

\noindent%
\begin{minipage}{\linewidth}
\makebox[\linewidth]{
  \includegraphics[keepaspectratio=true,scale=0.4]{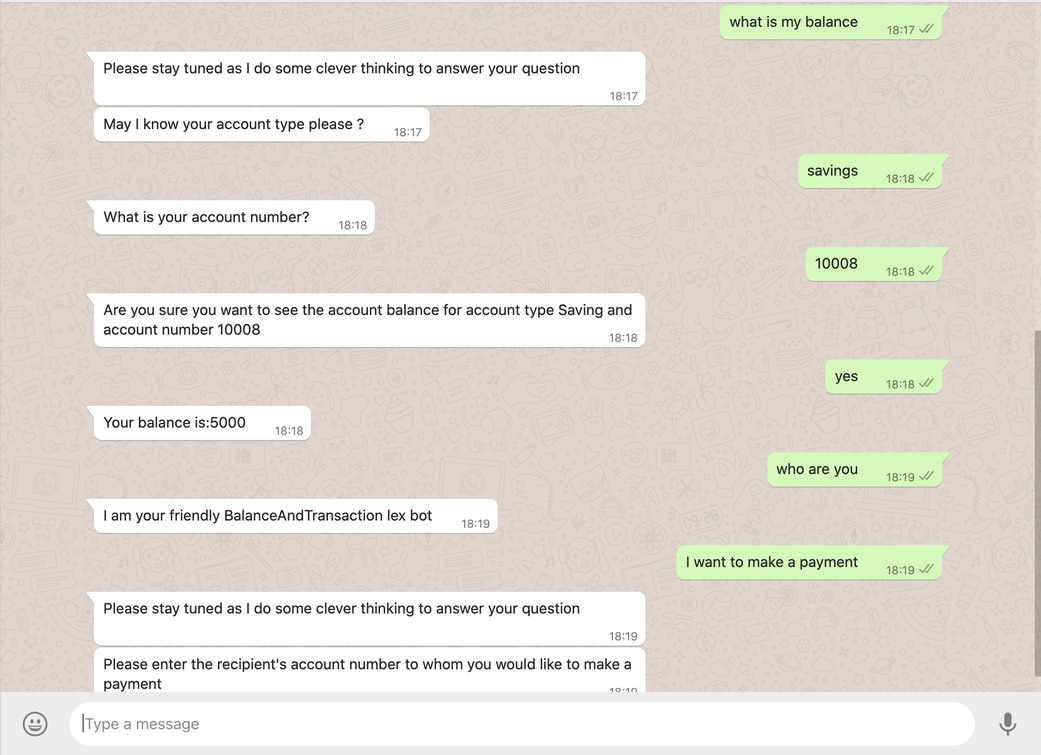}} 
 
\end{minipage}

\bigskip

\noindent%
\begin{minipage}{\linewidth}
\makebox[\linewidth]{
  \includegraphics[keepaspectratio=true,scale=0.4]{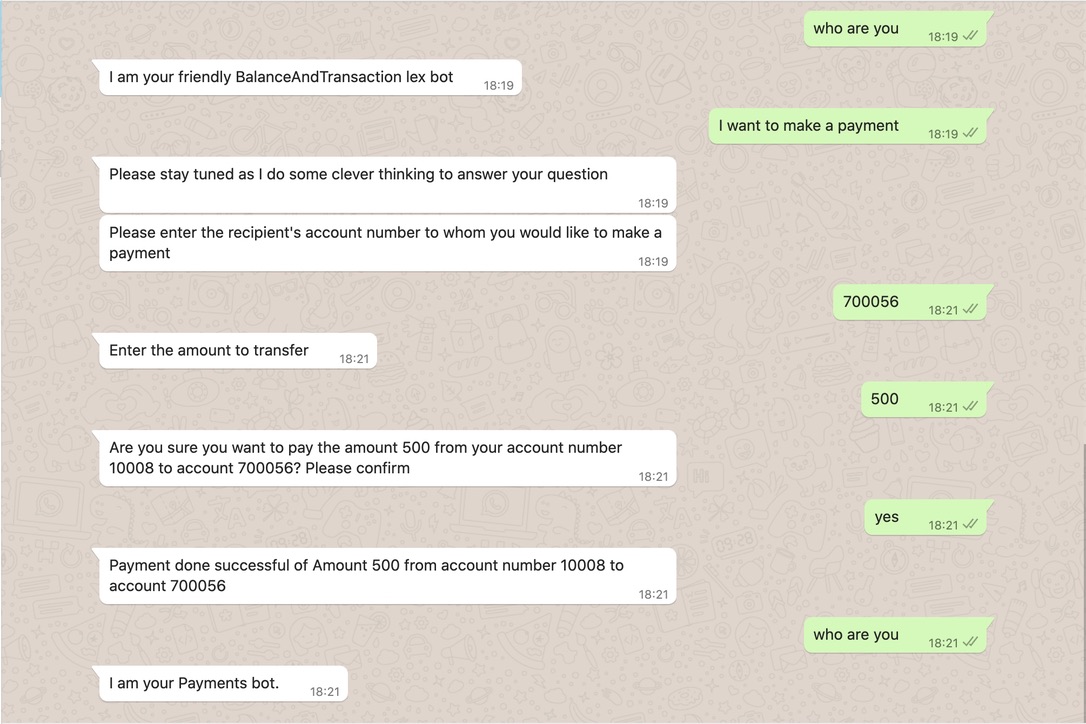}} 
 
\end{minipage}

\bigskip

\noindent%
\begin{minipage}{\linewidth}
\makebox[\linewidth]{
  \includegraphics[keepaspectratio=true,scale=0.4]{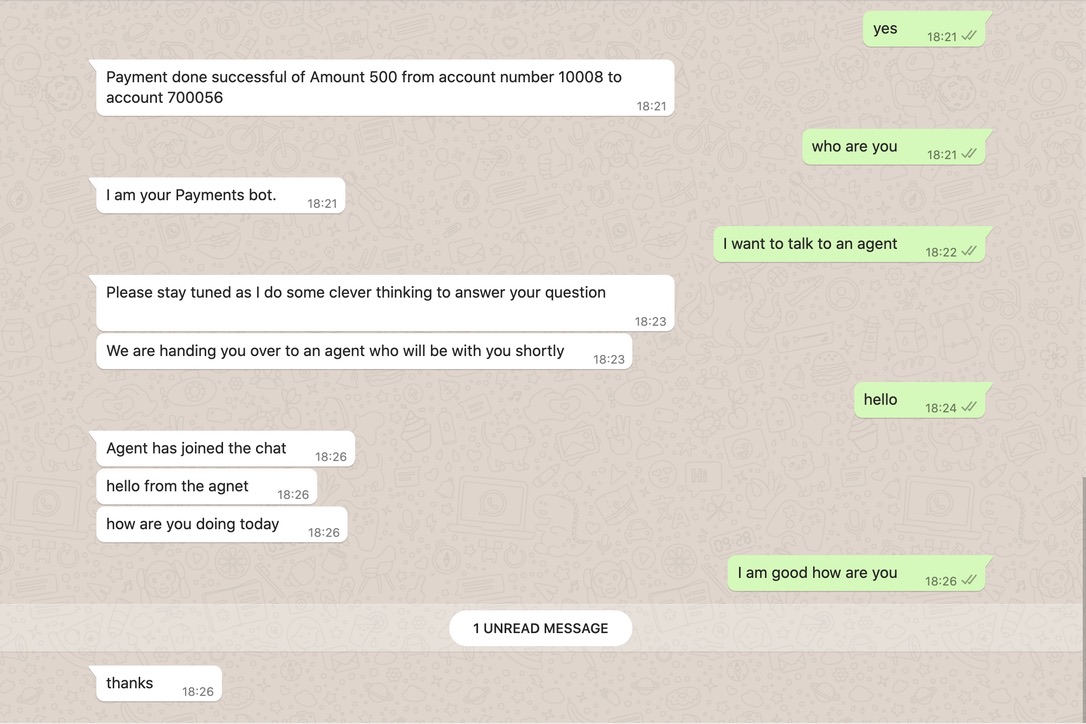}} 
 
\end{minipage}

\section{Conclusion}

We presented LPar platform that allows multiple distributed agents to seamlessly work together and deliver a seemless customer experience for Natural Language Interfaces. We also presented two strategies for Agent selection. We also established how these strategies are different from the Master Agent strategy that is currently utilized in most multi agent systems. With created an experimental setup of Banking domain and showcased how multiple bots of different categories and built over different vendor platforms worked together seamlessly.

\section{Future Work}

We would like to continue our research in area of Agent allocation and Context retention. We would like to represent the system as a Partially Observable Markov Decision Process and look to leverage advances in Reinforcement Learning to model the state switching.

\section{Acknowledgements}

I would like to thank following people as without their help this paper would not have net seen light of the day

\begin{itemize}
  \item Rosy Bharadwaj and Swati Rani for helping create the experimental setup. Really appreciate their efforts and patience for numerous design changes that I kept coming back with. Also, they solved many technical challenges  
  \item Shakeel Khan and Adrian Gubby for always being supportive of crazy ideas which in early days look far from realms of possiblity.  
  \item Salil Haridas and Amar Aggarwal for being my mentors at Cognizant.
  \item My family, as most of this documentation was done at a time which ideally should be given to them
  
  \end{itemize}

\bibliographystyle{unsrt}  


\begin{thebibliography}{1}



\bibitem{hadash2018estimate}
\newblock Akari Asai, Kazuma Hashimoto, Hannaneh Hajishirzi, Richard Socher, and Caiming Xiong.
Learning to retrieve reasoning paths over wikipedia graph for question answering. In ICLR
2020, 2020


\bibitem{brt}
\newblock Nils Reimers, Iryna Gurevych: Sentence-BERT: Sentence Embeddings using Siamese BERT-Networks

\bibitem{bbhdj}
\newblock Babak Hodjat, Makoto Amamiya: Adaptive Interaction Using the Adaptive Agent Oriented Software
Architecture 


\end{thebibliography}

\end{document}